\documentclass{article}
\usepackage{spconf,amsmath,graphicx,hyperref}

\usepackage{amssymb}
\usepackage{amsmath}

\usepackage{makecell}
\usepackage{color}
\usepackage{multirow}
\usepackage{times}
\usepackage{soul}
\usepackage{url}

\usepackage[small]{caption}
\usepackage{graphicx}
\usepackage{amsmath}
\usepackage{amsthm}
\usepackage{booktabs}
\usepackage{algorithm}
\usepackage{algorithmic}

\newcommand{\hanxiR}{\mathbb{R}}
\newcommand{\II}{\text{I}}

\title{Towards Efficient Pixel Labeling for Industrial Anomaly Detection and Localization}
\name{Jingqi Wu$^{1,2}$ \qquad  Hanxi Li$^{1}$\sthanks{Hanxi Li is the corresponding author. This project was partially supported by National Natural Science Foundation of China 62372150.} \qquad  Lin Wu$^{3}$ \qquad   Hao Chen$^{4}$ \qquad  Deyin Liu$^{5}$ \qquad  Peng Wang$^{6}$}
\address{
  $^1$Jiangxi Normal University, Jiangxi, China \\
  $^2$Southern University of Science and Technology, GuangDong, China\\
  $^3$Swansea University, Swansea, United Kingdom\\
  $^4$Zhejiang University, Zhejiang, China\\
  $^5$Anhui University, Jiangsu, China\\
  $^6$Northwestern Polytechnical University, Shanxi, China
  }
\begin{document}
%
\maketitle
\begin{abstract}

  
Industrial product inspection is often performed using Anomaly Detection (AD) frameworks trained solely on non-defective samples. Although defective samples can be collected during production, leveraging them usually requires pixel-level annotations, limiting scalability. To address this, we propose \textbf{ADClick}, an Interactive Image Segmentation (IIS) algorithm for industrial anomaly detection. ADClick generates pixel-wise anomaly annotations from only a few user clicks and a brief textual description, enabling precise and efficient labeling that significantly improves AD model performance (e.g., AP = 96.1\% on MVTec AD). 

We further introduce \textbf{ADClick-Seg}, a cross-modal framework that aligns visual features and textual prompts via a prototype-based approach for anomaly detection and localization. By combining pixel-level priors with language-guided cues, ADClick-Seg achieves state-of-the-art results on the challenging ``Multi-class'' AD task (AP = 80.0\%, PRO = 97.5\%, Pixel-AUROC = 99.1\% on MVTec AD).

\end{abstract}




\begin{keywords}
Anomaly Detection, Interactive Image Segmentation, Multi-Modal Fusion



\end{keywords}

\section{Introduction}
\label{sec:intro}
Automatic Optical Inspection (AOI) is essential in modern manufacturing. Most existing research adopts Anomaly Detection (AD) methods that avoid anomalous samples during training—not because such samples are unavailable, but because pixel-level annotation is prohibitively time-consuming. Since manual labeling is the primary bottleneck, models struggle to adapt quickly to newly observed defects.

Some works generate synthetic anomalies for training~\cite{zhang2023destseg,Liu_2023_CVPR}, while others show that real defect samples can improve accuracy~\cite{li2023efficient,li_target_2023}, though they still rely on costly pixel-level annotations. Recent studies attempt to reduce labeling costs with weak labels (e.g., bounding boxes)~\cite{li2023efficient}, but the low precision leads to degraded performance. 

In this paper, we propose \textbf{ADClick}, 
a semi-automatic labeling tool that efficiently produces high-quality pixel-wise anomaly labels from a few user clicks and a short defect description. ADClick integrates multi-modal cues---image features, language prompts, and residual features---to generate dense anomaly masks at a cost comparable to weak labels. For example, annotating a scratch on a metal nut requires only 2--5 clicks plus a brief textual prompt.

We further introduce \textbf{ADClick-Seg}, which embeds our multi-modal feature fusion module into anomaly detection, achieving state-of-the-art results under the challenging ``Multi-class'' setting. Extensive experiments confirm the effectiveness of our approach.

\textbf{Contributions:}
\begin{itemize}
    \item We pioneer the use of Interactive Image Segmentation (IIS) for efficient anomaly labeling, enabling pixel-wise annotations at weak-label cost.
    \item We enhance annotation quality via fusion of residual and linguistic features, yielding SOTA anomaly label generation.
    \item We propose ADClick-Seg, a multi-modal detection framework that achieves new benchmarks in anomaly detection and localization.
\end{itemize}


\section{Method}
\label{sec:method}

\begin{figure*}[t!] \centering{ \includegraphics [width=0.95\textwidth]{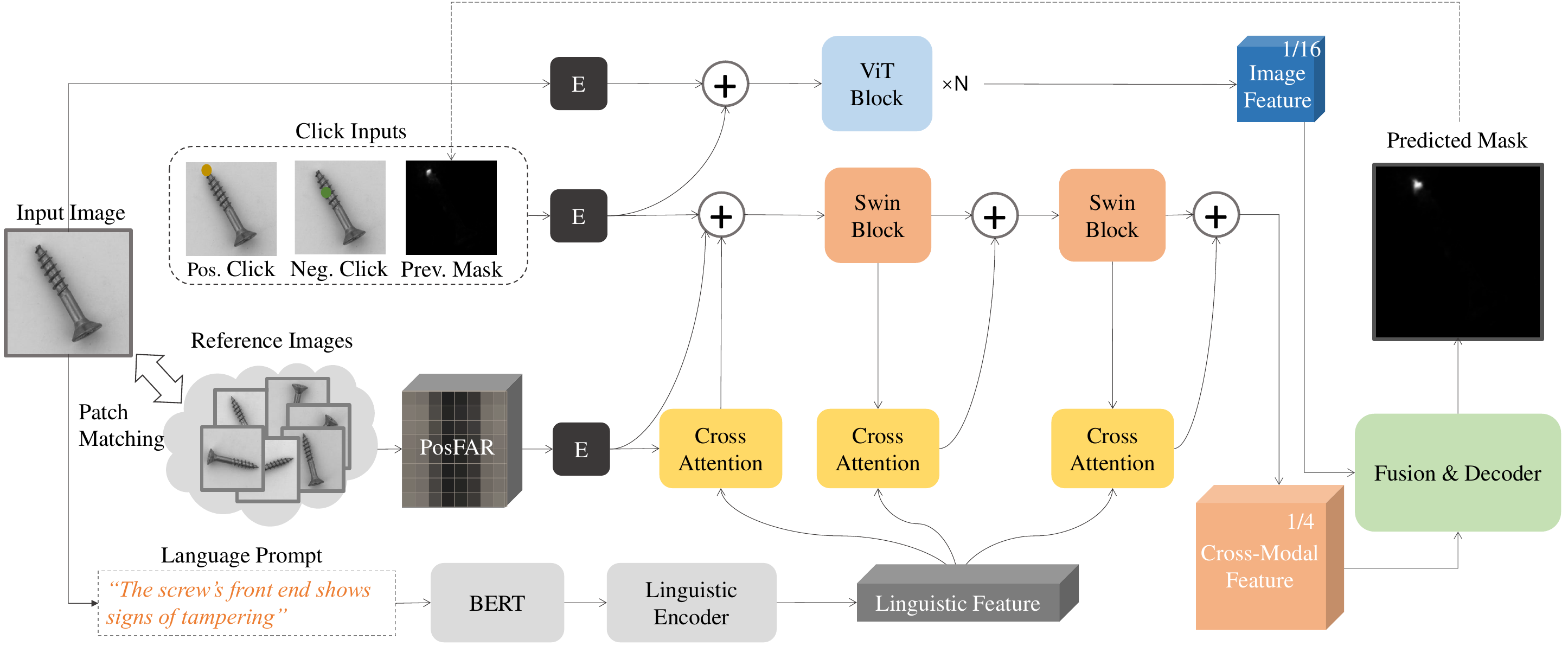} \caption{ The overview of the proposed ADClick. The model takes four input sources: the query image, the reference (defect-free) images, the language guidance, and the manual clicks. The dark-gray boxes labeled as `E'' represent the linear embedding process.
Better viewed in color. } \label{fig:method} } \end{figure*}

\subsection{Overview}
\label{subsec:structure}

Recent raw-pixel-based Interactive Image Segmentation (IIS) methods have shown strong applicability in real-world scenarios~\cite{liu_simpleclick_2023,huang2024focsam}.
Inspired by residual-based anomaly detection~\cite{roth2022towards,li2023efficient}, we extend this paradigm to interactive anomaly labeling by leveraging residual features, which are inherently more robust to noise and generalize better.  

Fig.~\ref{fig:method} illustrates the architecture of ADClick. The model takes four inputs: the query image, reference (defect-free) images, language guidance, and user clicks. Residual features, termed ``PosFAR'' from WeakREST~\cite{li2023efficient}, are fused with click embeddings and processed by a Swin Transformer backbone, where defect-specific linguistic features guide learning via cross-attention. In parallel, a ViT-based branch integrates clicks with raw-pixel information. The outputs of both branches are combined in a ``Fusion \& Decoder'' module to predict dense anomaly masks, which are iteratively refined across clicks.  

Formally, given a normalized input image $\II_{\text{tst}} \in \mathbb{R}^{H_I \times W_I \times 3}$ and a sequence of clicks $\mathcal{C} = \{{\bf C}_t = [x_t, y_t, \beta_t]^\text{T}\}$ with $\beta_t \in \{0,1\}$ indicating positive (anomalous) or negative (normal) clicks, the anomaly mask $\mathrm{M}_t$ at iteration $t$ is updated as:
\begin{equation}
  \mathrm{M}_t = \Phi_{\text{Click}}(\mathfrak{R}, \mathrm{M}_{t-1}, {\bf C}_t, \mathbf{l}, \II_{\text{tst}}),
\end{equation}
where $\mathfrak{R} \in \mathbb{R}^{h_f \times w_f \times d_f}$ denotes stored PosFAR features, $\mathbf{l} \in \mathbb{R}^Z$ is the linguistic feature of the task description, and $\mathrm{M}_{t-1}$ is the mask from the previous iteration.

\subsection{Interactive Segmentation with Positional Fast Anomaly Residuals}
\label{subsec:location_aware}

We adopt WeakREST~\cite{li2023efficient} to generate residual features, termed
``PosFAR,'' which jointly capture global and local similarities between the test image and
the reference bank. Specifically, we first extract a set of deep feature vectors from the
test image $\II_{\text{tst}}$:
\begin{equation}
   \{\mathbf{p}^1_{\text{tst}}, \mathbf{p}^2_{\text{tst}}, \cdots,
   \mathbf{p}^M_{\text{tst}}\} = \Psi_{\mathrm{PCF}}(\II_{\text{tst}}),
\end{equation}
where each $\mathbf{p}^j_{\text{tst}} \in \mathbb{R}^{d_f}$ represents a
\textit{Position Constrained Feature} (PCF)~\cite{li2023efficient} for the $j$-th patch.
In the same way, anomaly-free references are extracted as
$\mathcal{P}_{\text{ref}} = \{\mathbf{p}^1_{\text{ref}}, \mathbf{p}^2_{\text{ref}}, \cdots, \mathbf{p}^N_{\text{ref}}\}$.  

For each patch feature $\mathbf{p}^j_{\text{tst}}$, its
\textit{Positional Fast Anomaly Residual} (PosFAR) is computed as:
\begin{equation}
  \mathbf{r}_j = \lceil \text{ABS}(\mathbf{p}^j_{\text{tst}} -
  \mathbf{p}^{\ast}_{\text{ref}}) \rceil^{\theta} \in \mathbb{R}^{d_f}, 
  \quad j=1,2,\dots,M,
\end{equation}
where $\text{ABS}(\cdot)$ denotes the element-wise absolute value, $\lceil \cdot
\rceil^{\theta}$ is the element-wise $\theta$-power operation, and
$\mathbf{p}^{\ast}_{\text{ref}}$ is the matched reference feature selected from
$\mathcal{P}_{\text{ref}}$ using the WeakREST strategy.  


\begin{figure}[ht]  
\centering
{
\includegraphics [width=0.35\textwidth]{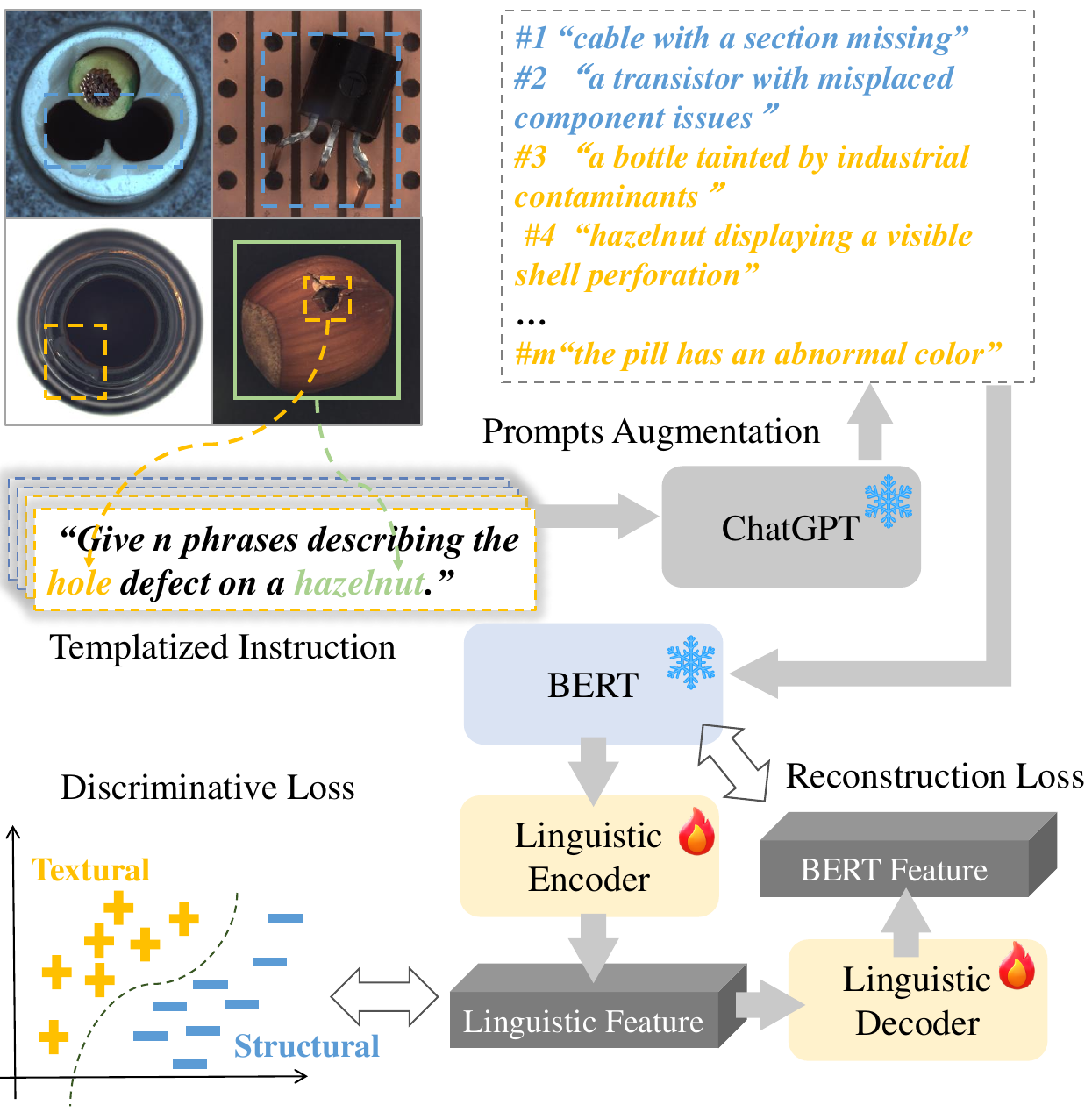}
\caption{
The training of defect-specific linguistic features follows a simple yet effective workflow. Users only need to provide keywords to form templatized instructions, from which ChatGPT generates descriptive phrases. 
}
  \label{fig:language}
}
\end{figure}
\subsection{Discriminative Linguistic Features}
\label{subsec:language}

Inspired by referring image segmentation~\cite{yang2022lavt}, 
we introduce defect-specific linguistic features to provide richer semantic cues beyond binary ``defective/non-defective'' labels, improving detection quality.

As illustrated in Fig.~\ref{fig:language}, candidate prompts are first generated using
ChatGPT~\cite{wu2023brief} with a templatized instruction, e.g.,
\begin{equation}
  \resizebox{.7\hsize}{!}{$
  \text{``Give } U \text{ phrases describing the } \{def\} \text{ defect on a } \{obj\}."$}
\end{equation}
where $def$ and $obj$ denote defect and object categories, and $U$ controls the number of
variants. Each prompt $\phi$ is encoded with BERT~\cite{devlin2018bert} and further mapped
into a compact linguistic representation $\mathbf{l}$ via a trainable encoder:
\begin{equation}
  \phi \xrightarrow{\Psi_{\text{BERT}}} \mathbf{v} \xrightarrow{\Psi_{\text{En}}} \mathbf{l}
  \in \mathbb{R}^{Z}.
\end{equation}


During inference, a randomly selected ChatGPT-generated description is used as input. The
resulting linguistic feature $\mathbf{l}$ is fused with visual residual features
$\mathbf{R}_{l}$ via cross-attention~\cite{vaswani2017attention}:
\begin{equation}
\label{eq:cross_attention}
  \mathbf{R}_{L}^{'} =
  \text{softmax}\!\left(\tfrac{\Psi_{\text{Conv}}(\mathbf{R}_{l})^T \mathbf{W}_K \mathbf{l}}
  {\sqrt{d_f}}\right)\mathbf{W}_V \mathbf{l},
\end{equation}
where $\Psi_{\text{Conv}}(\cdot)$ is a $1\times1$ convolution with normalization, and
$\mathbf{W}_K$, $\mathbf{W}_V$ are projection matrices. This integration injects
fine-grained defect semantics into the AD process.

\subsection{Dense Label Estimation}
We design a raw-pixel branch to complement residual features for dense label estimation.
The proposed Fusion \& Decoder module integrates cross-modal and image features through element-wise addition across multiple scales ($\frac{1}{32}$–$\frac{1}{4}$ of input resolution). Before fusion, cross-modal tensors are adapted using Zero-Convolution (ZC) layers \cite{zhang2023adding}, effectively functioning as a ControlNet \cite{zhang2023adding} for the raw-pixel branch. The fusion process is formalized as:
\begin{equation}
    \label{equ:zero_conv}
  \begin{split} 
 \mathbf{F}^{img}_{i} & = \Psi^{img}_{\text{ConvNeck}_i}(\mathbf{F}_N) \in \hanxiR^{d_i \times h_i \times w_i} ~\forall i, \\ 
     \mathbf{F}^{res}_{i} & = \Psi^{res}_{\text{ConvNeck}_i}(\mathbf{R}_{L}^{'}) \in \hanxiR^{d_i \times h_i \times w_i}  ~\forall i, \\       
  \mathbf{F}^{fus}_{i} & =  \mathbf{F}^{img}_{i}  + \Psi_{\text{ZC}_i}(\mathbf{F}^{res}_{i}) \in \hanxiR^{d_i \times h_i \times w_i}  ~\forall i, \\    
  \end{split} 
\end{equation}
where $\mathbf{F}_N$ is the output of the ViT image branch and $\mathbf{R}_{L}^{'}$ is the residual feature fused with linguistic cues (Eq.\ref{eq:cross_attention}). $\Psi_{\text{ConvNeck}i}$ extracts features at different scales, while $\Psi_{\text{ZC}_i}$ adapts residual features for stable fusion. Multi-scale fused features ${\mathbf{F}^{fus}_i}$ are then aggregated and decoded into dense label predictions. 

\subsection{Cross-modal Anomaly Detection and Localization}
We further enhance ADClick—termed \textbf{ADClick-Seg}—to achieve strong performance in both anomaly detection and localization. Since fully automatic segmentation models are more prone to overfitting on smaller datasets than human-in-the-loop models, we simplify the Fusion \& Decoder module and reassign the Zero-Convolution (ZC) modules to the image branch, improving stability and reducing overfitting.



At test time, since the defect type is unknown, we perform inference over all $P$ defect types and store predicted maps $\{\mathrm{A}_i \in \mathbb{R}^{H_I \times W_I} \mid i=1,2,\dots,P\}$. The final anomaly score map is computed as:

\begin{equation}
    \label{equ:final_score}
    \forall(x, y), ~~\hat{\mathrm{A}}[x, y] = \max_{i=1}^{P} \mathrm{A}_i[x, y].
\end{equation}


\subsection{Implementation Details}
\label{subsec:detail}

The image branch of ADClick is based on the SimpleClick framework~\cite{liu_simpleclick_2023}, with weights pre-trained on COCO~\cite{lin2014microsoft} and LVIS~\cite{gupta2019lvis} and trained using Normalized Focal Loss~\cite{sofiiuk2019adaptis}. Following WeakREST~\cite{li2023efficient}, the residual branch uses two Swin Transformer blocks, each with 32 heads for 512-dimensional residual features.


For input images of size $1024 \times 1024$, we use WideResNet-50~\cite{zagoruyko2016wide} pre-trained on ImageNet-1K as the feature extractor to obtain PosFAR~\cite{li2023efficient}. The Linguistic Encoder is trained on $14 \times 40$ prompt features, augmented and encoded using ChatGPT 3.5~\cite{wu2023brief} and BERT~\cite{devlin2018bert}, producing a $Z=512$-dimensional feature with $\lambda=0.05$. Training uses AdamW with a learning rate of $1e\!-\!5$ and weight decay of 0.05.

\section{Experiments}\label{sec:exp}

\subsection{Experimental Settings}
\label{subsec:setting}
We evaluate the proposed ADClick and ADClick-Seg against state-of-the-art Interactive Image Segmentation (IIS) and Anomaly Detection (AD) methods, respectively. IIS comparisons include SimpleClick~\cite{liu_simpleclick_2023}, GPCIS~\cite{zhou_interactive_2023}, and FocSAM~\cite{huang2024focsam}, while ADClick-Seg (trained in a “Multi-class’’ setting or with 3–5 clicks) is compared with leading AD approaches~\cite{Liu_2023_CVPR,zhang2023destseg,he2024mambaad,guo2024dinomaly,li2023efficient,roth2022towards, guo2024recontrast,ristea2022self}.

Experiments are conducted on MVTec-AD~\cite{bergmann2019mvtec} and KolektorSDD2~\cite{bovzivc2021mixed}, using Image-AUROC, Pixel-AUROC, PRO, and AP for AD, and mIoU and NoC@80 for IIS. All experiments run on an Intel i5-13600KF CPU, 64GB RAM, and an NVIDIA RTX 4090 GPU.
\begin{table*}[!htb]
	\centering
		\resizebox{\textwidth}{!}{
		\begin{tabular}{@{}l|c|cccc|cccc@{}}
			\toprule
            \multirow{2}{*}{Method}               & \multirow{2}{*}{Backbone}   &\multicolumn{4}{c|}{MVTec AD}&\multicolumn{4}{c}{KolektorSDD2}\\
            \cmidrule{3-10}
			&& 2-click & 3-click & 5-click & NoC80& 2-click & 3-click & 5-click & NoC80 \\ \midrule
			FocSAM$_\text{CVPR'24}$ \cite{huang2024focsam}        & ViT-H      & 37.3/62.5/85.3/57.0            & 51.9/71.1/88.4/67.9          & 73.9/80.6/92.1/78.5          & 6.2  & 40.5/72.1/87.2/65.1 & 72.1/80.2/90.3/73.8 & 83.3/87.6/97.5/{\color{red}{\textbf{82.3}}} & {\color{red}{\textbf{5.0}}} \\
			FocSAM$^{\star}_\text{CVPR'24}$ \cite{huang2024focsam}      & ViT-H      & 43.1/65.6/85.7/58.6          & 61.5/72.3/88.2/68.3          & 77.6/81.0/92.2/78.9            & 6.4 & 41.3/71.8/87.2/64.6 & 72.9/79.2/89.7/73.0 & 83.4/86.5/93.8/81.6 & 5.3  \\
			GPCIS$_\text{CVPR'23}$ \cite{zhou_interactive_2023}   & ResNet50   & 40.5/70.9/90.6/32.2          & 54.4/77.4/92.9/42.7          & 75.0/85.1/96.1/58.1            & 10.0 & 62.6/88.8/94.3/54.2 & 79.7/94.2/96.8/62.8 & 91.3/95.3/98.3/72.3 & 7.8   \\
			GPCIS$^{\star}_\text{CVPR'23}$ \cite{zhou_interactive_2023}       & ResNet50   & 58.8/79.9/87.4/49.0            & 71.1/85.2/93.0/57.5            & 84.6/90.9/97.6/68.7          & 9.2     & 65.4/89.5/92.2/52.3 & 72.6/91.0/95.4/58.6 & 82.0/94.4/98.0/66.3   & 9.3   \\
			SimpleClick$_\text{ICCV'23}$ \cite{liu_simpleclick_2023}  & ViT-B      & 38.2/70.7/91.4/41.9          & 58.5/83.0/95.1/53.0              & 80.4/91.9/97.7/71.5          & 6.8   & 8.5/46.4/83.1/51.0  & 21.6/79.7/92.7/59.8 & 73.8/94.5/96.7/72.0 & 6.3  \\
			SimpleClick$^{\star}_\text{ICCV'23}$ \cite{liu_simpleclick_2023}  & ViT-B      & 69.4/90.1/95.4/54.5          & 78.6/92.7/97.2/61.5          & 87.0/95.5/98.3/72.7            & 6.8   & 89.1/97.2/97.6/67.8 & 92.3/97.4/98.4/73.2 & 96.2/99.0/99.3/80.5   & 5.4\\
			ADClick $^{\star}$                  & ViT-B+Swin-M & {\color{red}{\textbf{90.1}}}/{\color{red}{\textbf{95.9}}}/{\color{red}{\textbf{99.0}}}/{\color{red}{\textbf{69.2}}}            &  {\color{red}{\textbf{93.3}}}/{\color{red}{\textbf{97.2}}}/{\color{red}{\textbf{99.4}}}/{\color{red}{\textbf{75.0}}}            &  {\color{red}{\textbf{96.1}}}/{\color{red}{\textbf{98.3}}}/{\color{red}{\textbf{99.7}}}/{\color{red}{\textbf{81.1}}}          &  {\color{red}{\textbf{5.6}}}{\color{red}{\textbf{91.5}}}/{\color{red}{\textbf{98.0}}}/{\color{red}{\textbf{98.4}}}/{\color{red}{\textbf{72.0}}}     & {\color{red}{\textbf{95.5}}}/{\color{red}{\textbf{98.4}}}/{\color{red}{\textbf{99.5}}}/{\color{red}{\textbf{76.2}}} & {\color{red}{\textbf{97.7}}}/{\color{red}{\textbf{99.3}}}/{\color{red}{\textbf{99.9}}}/81.0 & 5.3   \\ \bottomrule
		\end{tabular}
	}
  \caption{The comparison on the AP, PRO, Pixel AUROC, mIoU and NoC80 metrics for
  interactive segmentation on the MVTec AD dataset and KolektorSDD2 dataset.  $^{\star}$ denotes a model finetuned
  with all the subcategories of the MVTec AD dataset except the one currently under
  test. Note that here ``Swin-M'' indicates ``Swin-Micro'' which contains only $2$
  Swin-transformer blocks.}
	\label{table:iis_mvtec_result}
\end{table*}

\subsection{Ablation Study}
\label{subsec:ablation}

\begin{table*}[!t]
	\centering

	\resizebox{0.8\linewidth}{!}{
		\begin{tabular}{@{}cccccccc@{}}
			\toprule
			Image Feature             & Residual Feature          & Language                  & Zero conv.                & 2-click             & 3-click             & 5-click             & NoC80 \\ \midrule
			\checkmark &                           &                           &                           & 69.4/90.1/95.4/54.5 & 78.6/92.7/97.2/61.5 & 87.0/95.5/98.3/72.7   & 6.8   \\
			& \checkmark &                           &                           & 80.0/93.1/97.0/54.6     & 82.7/94.1/97.6/59.2 & 86.4/95.2/98.3/65.2 & 11.6  \\
			& \checkmark & \checkmark &                           & 80.3/92.9/97.0/55.4   & 83.5/94.2/97.7/60.4 & 87.6/95.4/98.5/66.7 & 11.3  \\
			\checkmark & \checkmark & \checkmark &                           & 87.7/94.6/98.3/68.8 & 92.3/96.0/99.0/74.5     & 95.6/98.2/99.6/80.4 & 5.7   \\
			\checkmark & \checkmark & \checkmark & \checkmark & {\color{red}{\textbf{90.1}}}/{\color{red}{\textbf{95.9}}}/{\color{red}{\textbf{99.0}}}/{\color{red}{\textbf{69.2}}}   & {\color{red}{\textbf{93.3}}}/{\color{red}{\textbf{97.2}}}/{\color{red}{\textbf{99.4}}}/{\color{red}{\textbf{75.0}}}   & {\color{red}{\textbf{96.1}}}/{\color{red}{\textbf{98.3}}}/{\color{red}{\textbf{99.7}}}/{\color{red}{\textbf{81.1}}} & {\color{red}{\textbf{5.6}}}   \\ \bottomrule
		\end{tabular}
	}
    	\caption{
		Ablation study results (AP/PRO/P\_AUROC/mIoU and NoC80) of ADClick on MVTec-AD.
	}	
	\label{table:ablation_click_result}
\end{table*}

To analyze module contributions, we conduct an ablation study for ADClick, summarized in Tab.~\ref{table:ablation_click_result}. The results show that image information alone is insufficient for accurate labeling. Using residual features significantly improves performance over raw-pixel input for 2- and 3-click cases, though this advantage reverses with 5 clicks. Language guidance further boosts accuracy, and the zero-convolution module, which fuses image and cross-modal features, clearly contributes to improved performance.

\subsection{Label Generation Accuracy}
\label{subsec:label}

To evaluate the proposed label generation strategy, we establish supervision settings where off-the-shelf IIS methods and our ADClick model are pre-trained on all subcategories of the MVTec AD dataset except the one under test. For KolektorSDD2, ADClick is pre-trained on all texture categories of MVTec AD.

As shown in Tab.\ref{table:iis_mvtec_result}, ADClick consistently outperforms state-of-the-art IIS methods across all evaluation metrics on MVTec AD, regardless of the number of clicks (2, 3, or 5). On KolektorSDD2, similar trends are observed, with the exception that FocSAM~\cite{huang2024focsam} slightly surpasses ADClick in mIoU with 5 clicks and in NoC@80. These results highlight the effectiveness of ADClick in generating dense labels from sparse click inputs.

\subsection{Anomaly Detection and Localization}
\label{subsec:ad_performance}

To evaluate the proposed fusion of image (Img) and discriminative linguistic (Lang) features, we compare ADClick-Seg with SOTA anomaly detection methods under the ``Multi-class'' setting, where a single model is trained on defect-free images across all subcategories. Tab.~\ref{tab:seg_mvtec}  shows that ADClick-Seg outperforms WeakRest and most SOTA methods, with only a slight ($0.7\%$) gap to Dinomaly~\cite{guo2024dinomaly} in Image-AUROC on MVTec-AD. Tab.~\ref{table:ksdd2_seg_result} shows results on KolektorSDD2, where linguistic features are not used since the dataset contains a single object type without defect categories.

\begin{table}[!htb]
\centering
\resizebox{\linewidth}{!}{
\begin{tabular}{@{}lccccc@{}}
\toprule
Method& Supervision                & AP   & PRO  & P-AUROC & I-AUROC \\ \midrule
SimpleNet \cite{Liu_2023_CVPR}&Un&45.9&86.5&96.8&95.3\\
DeSTSeg \cite{zhang2023destseg}&Un&54.3&64.8&93.1&89.2\\
ReContrast \cite{guo2024recontrast}&Un&60.2&93.2&97.1&98.3\\
MambaAD \cite{he2024mambaad}&Un&56.3&93.1&97.7&98.6\\
Dinomaly \cite{guo2024dinomaly}&Un&69.3&94.8&98.4&{\color{red}{\textbf{99.6}}}\\
WeakREST  \cite{li2023efficient} &Un    &79.0&97.3&{\color{red}{\textbf{99.1}}}&98.8 \\
 ADClick-Seg (Res+Lang) & Un &79.5&97.4&{\color{red}{\textbf{99.1}}}&99.0\\
 ADClick-Seg (Res+Lang+Img)  & Un &{\color{red}{\textbf{80.0}}}&{\color{red}{\textbf{97.5}}}&{\color{red}{\textbf{99.1}}}&99.0\\ \midrule
 WeakREST  \cite{li2023efficient} &GT    &{\color{red}{\textbf{84.3}}}&97.8&{\color{red}{\textbf{99.5}}}&99.3 \\
 WeakREST  \cite{li2023efficient} & ADClick-5   &83.8&{\color{red}{\textbf{98.0}}}&{\color{red}{\textbf{99.5}}}&{\color{red}{\textbf{99.4}}} \\
 WeakREST \cite{li2023efficient} & ADClick-3   &83.2&97.8&{\color{red}{\textbf{99.5}}}&99.3 \\
\bottomrule
\end{tabular}
}
\caption{
 Performance on MVTec AD under ``Multi-class'' setting. 
 Note that the upper sub-table shows the results obtained in the
  unsupervised condition and the lower part reports those with genuine defective samples. The supervision method ``ADClick-$k$'' indicates the pseudo-label generated by using ADClick with $k$ clicks.   
  }
\label{tab:seg_mvtec}
\end{table}

\begin{table}[!thb]
	\centering
	\resizebox{\linewidth}{!}{
		\begin{tabular}{@{}lcccc@{}}\toprule
			Method   & AP   & PRO  & P\_AUROC & I\_AUROC \\\midrule
			PatchCore \cite{roth2022towards}   & 64.1& 88.8 & 97.1       & 94.6  \\
			SSPCAB \cite{ristea2022self}               & 44.5 & 66.1 & 86.2   & 83.4      \\
			RD \cite{deng2022anomaly}                  & 43.5 &94.7& 97.6   & 96.0   \\
			WeakREST \cite{li2023efficient} &76.4 & {\color{red}{\textbf{98.1}}} & {\color{red}{\textbf{99.6}}} & 96.8        \\
			ADClick-Seg (Res+Img)& {\color{red}{\textbf{77.0}}}& {\color{red}{\textbf{98.1}}} & {\color{red}{\textbf{99.6}}} & {\color{red}{\textbf{97.0}}}\\\bottomrule
		\end{tabular}
}
	\caption{
		Unsupervised anomaly localization and detection performance (AP, PRO, Pixel AUROC and
		Image AUROC) on KolektorSDD2. Note that no linguistic feature is used here due to the limitation of this dataset.}
	\label{table:ksdd2_seg_result}
\end{table}

To validate ADClick in practice, we conduct a ``supervised'' multi-class anomaly detection experiment. WeakREST is trained using ADClick-generated labels (3- or 5-click) and compared with training on ground-truth labels (GT). Results in the bottom of Tab.~\ref{tab:seg_mvtec} show that pseudo labels provide comparable information to GT, with 5 clicks sufficient for high-quality annotations.


	


\section{Conclusion}
\label{sec:conclusion}

This work addresses anomaly labeling to reduce the cost of manual pixel annotations in AD. We propose a transformer-based interactive segmentation model that generates high-quality pixel labels for anomalies using cross-modal inputs, including user clicks, language prompts, and deep feature residuals. The information fusion process and defect-specific linguistic features further improve standard AD tasks. Our method achieves state-of-the-art performance in Multi-class AD, paving the way for efficient label generation and enhanced detection with minimal supervision.


\bibliographystyle{IEEEbib}
\bibliography{anomaly_detection,interactive_segmentation,others}

\begin{thebibliography}{10}

\bibitem{zhang2023destseg}
Xuan Zhang, Shiyu Li, Xi~Li, Ping Huang, Jiulong Shan, and Ting Chen,
\newblock ``Destseg: Segmentation guided denoising student-teacher for anomaly detection,''
\newblock in {\em CVPR}, 2023, pp. 3914--3923.

\bibitem{Liu_2023_CVPR}
Zhikang Liu, Yiming Zhou, Yuansheng Xu, and Zilei Wang,
\newblock ``Simplenet: A simple network for image anomaly detection and localization,''
\newblock in {\em CVPR}, 2023, pp. 20402--20411.

\bibitem{li2023efficient}
Hanxi Li, Jingqi Wu, Hao Chen, Mingwen Wang, and Chunhua Shen,
\newblock ``Efficient anomaly detection with budget annotation using semi-supervised residual transformer,''
\newblock {\em arXiv e-prints}, 2023.

\bibitem{li_target_2023}
Hanxi Li, Jianfei Hu, Bo~Li, Hao Chen, Yongbin Zheng, and Chunhua Shen,
\newblock ``Target before {Shooting}: {Accurate} {Anomaly} {Detection} and {Localization} under {One} {Millisecond} via {Cascade} {Patch} {Retrieval},''
\newblock {\em IEEE TIP}, 2024.

\bibitem{liu_simpleclick_2023}
Qin Liu, Zhenlin Xu, Gedas Bertasius, and Marc Niethammer,
\newblock ``Simpleclick: Interactive image segmentation with simple vision transformers,''
\newblock in {\em ICCV}, 2023, pp. 22290--22300.

\bibitem{huang2024focsam}
You Huang, Zongyu Lan, Liujuan Cao, Xianming Lin, Shengchuan Zhang, Guannan Jiang, and Rongrong Ji,
\newblock ``Focsam: Delving deeply into focused objects in segmenting anything,''
\newblock in {\em CVPR}, 2024, pp. 3120--3130.

\bibitem{roth2022towards}
Karsten Roth, Latha Pemula, Joaquin Zepeda, Bernhard Sch\"olkopf, Thomas Brox, and Peter Gehler,
\newblock ``Towards total recall in industrial anomaly detection,''
\newblock in {\em CVPR}, 2022, pp. 14318--14328.

\bibitem{yang2022lavt}
Zhao Yang, Jiaqi Wang, Yansong Tang, Kai Chen, Hengshuang Zhao, and Philip~HS Torr,
\newblock ``Lavt: Language-aware vision transformer for referring image segmentation,''
\newblock in {\em CVPR}, 2022, pp. 18155--18165.

\bibitem{wu2023brief}
Tianyu Wu, Shizhu He, Jingping Liu, Siqi Sun, Kang Liu, Qing-Long Han, and Yang Tang,
\newblock ``A brief overview of chatgpt: The history, status quo and potential future development,''
\newblock {\em IEEE/CAA Journal of Automatica Sinica}, vol. 10, no. 5, pp. 1122--1136, 2023.

\bibitem{devlin2018bert}
Jacob Devlin Ming-Wei~Chang Kenton and Lee~Kristina Toutanova,
\newblock ``Bert: Pre-training of deep bidirectional transformers for language understanding,''
\newblock in {\em Proceedings of NAACL-HLT}, 2019, pp. 4171--4186.

\bibitem{vaswani2017attention}
Ashish Vaswani,
\newblock ``Attention is all you need,''
\newblock {\em arXiv preprint arXiv:1706.03762}, 2017.

\bibitem{zhang2023adding}
Lvmin Zhang, Anyi Rao, and Maneesh Agrawala,
\newblock ``Adding conditional control to text-to-image diffusion models,''
\newblock in {\em ICCV}, 2023.

\bibitem{lin2014microsoft}
Tsung-Yi Lin, Michael Maire, Serge Belongie, James Hays, Pietro Perona, Deva Ramanan, Piotr Doll{\'a}r, and C~Lawrence Zitnick,
\newblock ``Microsoft coco: Common objects in context,''
\newblock in {\em ECCV}. Springer, 2014, pp. 740--755.

\bibitem{gupta2019lvis}
Agrim Gupta, Piotr Dollar, and Ross Girshick,
\newblock ``Lvis: A dataset for large vocabulary instance segmentation,''
\newblock in {\em CVPR}, 2019, pp. 5356--5364.

\bibitem{sofiiuk2019adaptis}
Konstantin Sofiiuk, Olga Barinova, and Anton Konushin,
\newblock ``Adaptis: Adaptive instance selection network,''
\newblock in {\em ICCV}, 2019, pp. 7355--7363.

\bibitem{zagoruyko2016wide}
Sergey Zagoruyko and Nikos Komodakis,
\newblock ``Wide residual networks,''
\newblock {\em arXiv preprint arXiv:1605.07146}, 2016.

\bibitem{zhou_interactive_2023}
Minghao Zhou, Hong Wang, Qian Zhao, Yuexiang Li, Yawen Huang, Deyu Meng, and Yefeng Zheng,
\newblock ``Interactive {Segmentation} as {Gaussian} {Process} {Classification},''
\newblock in {\em CVPR}, 2023, pp. 19488--19497.

\bibitem{he2024mambaad}
Haoyang He, Yuhu Bai, Jiangning Zhang, Qingdong He, Hongxu Chen, Zhenye Gan, Chengjie Wang, Xiangtai Li, Guanzhong Tian, and Lei Xie,
\newblock ``Mambaad: Exploring state space models for multi-class unsupervised anomaly detection,''
\newblock {\em arXiv e-prints}, 2024.

\bibitem{guo2024dinomaly}
Jia Guo, Shuai Lu, Weihang Zhang, and Huiqi Li,
\newblock ``Dinomaly: The less is more philosophy in multi-class unsupervised anomaly detection,''
\newblock {\em arXiv e-prints}, 2024.

\bibitem{guo2024recontrast}
Jia Guo, Lize Jia, Weihang Zhang, Huiqi Li, et~al.,
\newblock ``Recontrast: Domain-specific anomaly detection via contrastive reconstruction,''
\newblock {\em NeurIPS}, vol. 36, 2024.

\bibitem{ristea2022self}
Nicolae-Cătălin Ristea, Neelu Madan, Radu~Tudor Ionescu, Kamal Nasrollahi, Fahad~Shahbaz Khan, Thomas~B. Moeslund, and Mubarak Shah,
\newblock ``Self-supervised predictive convolutional attentive block for anomaly detection,''
\newblock in {\em CVPR}, 2022, pp. 13566--13576.

\bibitem{bergmann2019mvtec}
Paul Bergmann, Michael Fauser, David Sattlegger, and Carsten Steger,
\newblock ``Mvtec {AD} - {A} comprehensive real-world dataset for unsupervised anomaly detection,''
\newblock in {\em CVPR}, 2019, pp. 9592--9600.

\bibitem{bovzivc2021mixed}
Jakob Bo{\v{z}}i{\v{c}}, Domen Tabernik, and Danijel Sko{\v{c}}aj,
\newblock ``Mixed supervision for surface-defect detection: From weakly to fully supervised learning,''
\newblock {\em Computers in Industry}, vol. 129, pp. 103459, 2021.

\bibitem{deng2022anomaly}
Hanqiu Deng and Xingyu Li,
\newblock ``Anomaly detection via reverse distillation from one-class embedding,''
\newblock in {\em CVPR}, 2022, pp. 9727--9736.

\end{thebibliography}

\end{document}